># Risk-anticipatory autonomous driving strategies considering vehicles' weights, based on hierarchical deep reinforcement learning

*Di Chen, Hao Li, Zhicheng Jin, Huizhao Tu\* and Meixin Zhu\**

*Abstract*—**Autonomous vehicles (AVs) have the potential to prevent accidents caused by drivers' errors and reduce road traffic risks. Due to the nature of heavy vehicles, whose collisions cause more serious crashes, the weights of vehicles need to be considered when making driving strategies aimed at reducing the potential risks and their consequences in the context of autonomous driving. This study develops an autonomous driving strategy based on risk anticipation, considering the weights of surrounding vehicles and using hierarchical deep reinforcement learning. A risk indicator integrating surrounding vehicles' weights, based on the risk field theory, is proposed and incorporated into autonomous driving decisions. A hybrid action space is designed to allow for left lane changes, right lane changes and car-following, which enables AVs to act more freely and realistically whenever possible. To solve the above hybrid decision-making problem, a hierarchical proximal policy optimization (HPPO) algorithm with an attention mechanism (AT-HPPO) is developed, providing great advantages in maintaining stable performance with high robustness and generalization. An indicator, potential collision energy in conflicts (PCEC), is newly proposed to evaluate the performance of the developed AV driving strategy from the perspective of the consequences of potential accidents. The performance evaluation results in simulation and dataset demonstrate that our model provides driving strategies that reduce both the likelihood and consequences of potential accidents, at the same time maintaining driving efficiency. The developed method is especially meaningful for AVs driving on highways, where heavy vehicles make up a high proportion of the traffic.**

*Index Terms*—**Autonomous vehicles, Decision making, Driving risk, Driving safety, Reinforcement learning.**

This research is supported by the National Natural Science Foundation of China (NSFC, Grant No. 71971162, and Grant No. 52372339), and the Key Project of Science and Technology Commission of Shanghai Municipality (22dz1203400). *(Co-corresponding author: Huizhao Tu and Meixin Zhu).* Di Chen and Hao Li are co-first authors.

Di Chen, Zhicheng Jin are with Intelligent Transportation Thrust, Systems Hub, The Hong Kong University of Science and Technology (Guangzhou), Guangdong 511453, China (e-mail: dichen@hkust-gz.edu.cn; zhichengjin@hkust-gz.edu.cn).

Hao Li, Huizhao Tu are with Key Laboratory of Road and Traffic Engineering of the Ministry of Education, College of Transportation Engineering, Tongji University, Jiading District, Shanghai 201804, China (e-mail: haoliti@tongji.edu.cn, huizhaotu@tongji.edu.cn).

Meixin Zhu is with the Systems Hub at the Hong Kong University of Science and Technology (Guangzhou), the Civil and Environmental Engineering Department at the Hong Kong University of Science and Technology and Guangdong Provincial Key Lab of Integrated Communication, Sensing and Computation for Ubiquitous Internet of Things (e-mail: meixin@ust.hk).

## I. INTRODUCTION

ACCORDING to a report from the US National Highway Traffic Safety Administration (NHTSA), approximately 40,990 individuals lost their lives in motor vehicle accidents in 2023 [1]. Human errors are the leading cause of traffic accidents [2]. Autonomous vehicles (AVs) have the potential to significantly reduce the occurrence of traffic accidents by preventing those caused by human error [3]. During the fourth quarter, Tesla [4] recorded an average of one collision per 5.39 million miles driven by users of its Autopilot system, an automated driver assistance system, while non-Autopilot users experienced one crash per 1 million miles driven. Recent data from the NHTSA since 2023 indicated that a vehicle crash occurs in the U.S. approximately every 0.81 million miles traveled [1]. Although advancements in automotive technology hold promise for enhancing safety, they cannot eliminate accidents. Therefore, it is imperative to devise decision-making strategies for autonomous vehicles that enable them to perceive and respond to risks in complex traffic scenarios, thereby fostering safer road environments [5].

This challenge requires a thorough understanding of car-following and lane-changing (LC) behaviors [6]. LCs are categorized into mandatory lane changes (MLC) and discretionary lane changes (DLC). MLCs are executed due to external factors or regulations, such as lane merging or drops, while DLCs are optional and typically aim to improve driving conditions, such as seeking higher speeds or better visibility. DLC poses greater challenges than MLC, occurring at any location and involving inherent complexity and uncertainty [7, 8]. DLCs significantly contribute to road accidents on highways [9], and inappropriate DLCs can have detrimental effects on highway safety and efficiency, leading to traffic breakdowns, oscillations, or upstream queue propagation [6, 10, 11]. Therefore, investigating DLCs is imperative. Various studies on lane changing on highways have addressed this issue. Yu, et al. [12] optimize a multi-player dynamic game model by considering the states of surrounding vehicles to ensure the accurate execution of lane-changing decisions of AVs. Xiao, et al. [13] proposes a LC trajectory prediction model based on an encoder-decoder architecture to gain insight into the underlying motivations behind LC behavior. However, decision-making solely based on lane-changing behaviors does not fully meet the practical requirements of driving. What's more, Atzmon, et al. [14] observed that learning algorithms tend to overfit by placing undue emphasis on relatively minor factors. In contrast, human drivers excel at extracting essential information from complex environments, where the volume of critical data is often manageable. Consequently, attention mechanisms serve as effective tools for isolating vital information in decision making [15].

In an attempt to reduce potential safety risks, risk-anticipation-based driving strategies are increasingly being studied and proposed. Komol, et al. [16] present an early prediction framework for classifying drivers' intended intersection movements, considering vehicle position, speed, acceleration, and yaw rate. Noh [17] proposed a decision-making framework for AVs at road intersections and utilized the time to enter (TTE) in combination with Bayesian networks to evaluate the potential threats. Chib and Singh [18] compiled the common safety indicators used for the safety assessment of driving systems, including Time to Collision (TTC), Worst Time to Collision (WTTC), Time to React (TTR), Time Headway (THW), Deceleration to Safe Time (DST), Stopping Distance (SD), Crash Potential Index (CPI) and Conflict Index (CI). In existing studies on autonomous driving decisions, vehicle safety typically considers factors such as distance, speed, acceleration [6], and collision probability to estimate the likelihood of collisions, while often overlooking crash severity in decision-making and strategy evaluation processes. Yan, et al. [19] use velocity change as a metric for simulated collision severity. Vehicles of various types and weights result in different consequences in accidents [20, 21], with heavy vehicles particularly causing more severe outcomes. Newton's second law of motion, concerning the conservation of momentum, suggests that mitigating injuries in collisions involving heavy vehicles is significantly more challenging than in collisions with other passenger cars [22, 23]. Lyman and Braver [24] reported an annual average of 4,000 passenger deaths in large truck collisions in the US, with the trend showing a slight increase. As Zou, et al. [25] suggested, vehicle weight characteristics should be considered an essential factor in traffic safety risk assessment to enhance safety on roads with mixed traffic flow. Considering vehicle types and weights when designing driving strategies for AVs could help mitigate potentially serious accidents involving heavy vehicles. However, to our best knowledge, the influence of vehicle weights on AV decision-making has not yet been considered. Additionally, artificial potential fields can consolidate discrete events into a unified risk assessment field, providing superior spatial expressiveness compared to non-integrated risk indicators [26].

This study addresses the challenge of enabling AVs to navigate highways safely and efficiently in mixed traffic flows, with particular consideration for the weights of various vehicle types, notably heavy vehicles. The contributions of this study can be summarized as follows:

- A hierarchical actor-critic framework has been proposed to enable hybrid actions for AVs. A hierarchical proximal policy optimization (HPPO) algorithm with attention (AT-HPPO) is developed to decide the actions, including both discrete and continuous actions and incorporate an attention mechanism to accelerate convergence and enhance stability.
- A new quantitative anticipated driving risk (ADR) indicator based on the artificial potential field theory has been in-

troduced and integrated into the reward function of our decision-making framework. This metric, which accounts for the weights of surrounding vehicles, offers enhanced spatial expressiveness for evaluating the driving risk of AVs.
- To assess our proposed AV driving strategy, potential collision energy in conflicts (PCEC) is proposed as an evaluation indicator that can reflect the consequences of different vehicle weights in accidents.
- The performance of the proposed AT-HPPO algorithm has been evaluated by comparing it with state-of-the-art algorithms in the Simulation of Urban Mobility (SUMO) [27] simulator under various traffic densities. Further comparison with human driver trajectory data from the highD dataset [28] confirms the exceptional performance of the proposed algorithm in real-world highway scenarios, demonstrating its strong generalization and robustness.

These contributions can help AVs reduce the potentially serious risks of collisions with heavy vehicles, especially on highways that have a high proportion of heavy vehicles, thereby improving traffic safety in mixed traffic flows.

The paper is structured as follows. Section 2 provides a literature review of related research. Section 3 elaborates on the hierarchical deep learning methodology with hybrid action spaces for AVs and introduces the AT-HPPO algorithm including the attention mechanism, the definition of ADR indicator, the risk- and efficiency-based reward function, and algorithm training. Section 4 outlines the experimental settings on a highway with a high density of heavy vehicles, along with the evaluation metrics and comparison baselines. Section 5 presents a discussion of the comparison results in both simulation and real-world scenarios. Finally, conclusions are drawn in Section 6.

## II. LITERATURE REVIEW

In this section, reviews of the driving decision-making and driving risk assessment for AVs are provided, respectively.

### A. Decision-making methods

There are three primary categories of AV decision-making methods: rule-based, optimization-based, and learning-based. Rule-based (or physics-based [29]) approaches encompass all the scientific assumptions that pertain to the motion of vehicles or traffic flow, including microscopic, mesoscopic, and macroscopic traffic models. Rule-based methods usually have limited accuracy [30], poor generalization capability [31], an absence of adaptive updating [32], and require much expert knowledge for the establishment of rules [33]. Lacking consideration of the anthropomorphic nature of decision-planning models and dynamic environmental adaptability, they cannot be applied to well-defined situations. Optimization-based methods are often inspired by path planning in mobile robotics, e.g., $A^*$ [34] or $RRT$ [35]. However, a limitation of these methods is that some non-linear and non-convex motion constraints used in motion planning require large computational budgets or even fail to find the optimal solution when dealing with complex systems or problems [36-38]. Learning-based methods [39-41] are advanced learning models that imitate human intelligence, using techniques such as deep learning, reinforcement learning and other advanced machine learning methods. One limitation of deep learning methods is the requirement for labeled driving data, which can be a laborious and knowledge-intensive task. Collecting driving data on crash or near-crash scenarios of Avs is difficult. Deep generative methods can be utilized for generating long-tail or realistic scenarios, while they consistently encounter the challenge of mode collapse [42]. Using a DRL algorithm, an autonomous agent can learn through cost-effective trial and error without requiring human labeling or supervision. Training these models in virtual simulators allows for the implementation of DRL-based techniques in scenarios where crashes or near-crashes may occur, ultimately aiding autonomous vehicles in avoiding collisions [43]. DRL has demonstrated significant success in solving complex tasks [44]. Demonstrating significant potential in learning driving strategies eliminates the need for pre-defined rules or complex system modeling [45]. However, applying DRL to the real world is challenging because of the dynamic environment and extensive interaction with surrounding vehicles.

Lately, numerous efforts have been undertaken to investigate autonomous driving based on deep reinforcement learning (DRL). Liao, et al. [46] utilized the dueling deep Q-network (DDQN) to obtain a highway decision-making strategy. Muzahid, et al. [47] proposed a centralized multi-vehicle control strategy by reinforcement learning (RL) and compared soft actor-critic (SAC) and proximal policy optimization (PPO) algorithms. Han and Miao [48] designed an information-sharing-based multi-agent reinforcement learning framework for connected and automated vehicles (CAVs). Previous work has mostly focused on applying the DDQN [46, 49] or deep deterministic policy gradient (DDPG) [40, 50] to learn to drive, inheriting the simple greedy strategy from Q-learning by adding a small noise to a deterministic policy for exploration [51]. The relationship between the higher-level decision layer and the lower-level control layer of autonomous vehicles is typically direct [52]. In recent years, research has been conducted on hierarchical frameworks that incorporate both layers [40, 52-54]. Chen, et al. [40] built a hierarchical DDPG algorithm for learning driving strategies using images. Al-Sharman, et al. [53] and Du, et al. [52] both use RL for decision making and optimization-based methods for controlling. However, RL directly maximizes a task-level objective, which leads to more robust control performance in the presence of unmodeled dynamics [55]. As outlined in the literature review above, the methods of decision-making progress within the hierarchical framework can be further enhanced and optimized.

### B. Traffic safety risk assessment

Traffic safety risk assessment is of crucial importance for risk-anticipation-based AV driving strategies. Generally, two main approaches exist for risk assessment: deterministic [56, 57] and probabilistic [17, 50, 58, 59].

The deterministic approach is a binary prediction approach using specific indicators and then setting thresholds to determine whether a vehicle is at risk. It can include risk assessment approaches on a single dimension, using non-integrated indicators such as time to collision (TTC), time to brake (TTB), time headway (THW), etc., or take a multi-dimensional approach using an integrated indicator combining risks from the surroundings. In terms of the single-dimension deterministic approach, non-integrated indicators are utilized and the obtained safety metric values are compared with predefined thresholds for the indicators. Glaser, et al. [60] present a vehicle trajectory-planning algorithm using TTC to measure the likelihood of accidents for highly autonomous vehicles. Kim and Kum [61] developed a threat prediction algorithm that employs future trajectory predictions for surrounding vehicles to accurately forecast TTC over an extended time. Although these methods have the advantage of being computationally efficient, and can accurately assess threat risk in single-lane driving scenarios [56], their performance in multi-lane scenarios is generally unsatisfactory. In terms of a deterministic approach for an overall risk assessment, with an integrated indicator, field theory is adopted to describe the combined risks of complex scenarios considering all surrounding vehicles. In contrast to risk assessment methods that use indicators, risk field models usually consider the effect of vehicle weights. Lee and Kum [62] proposed a risk assessment module called the Predictive Occupancy Map (POM) to compute potential risks associated with surrounding vehicles based on their relative position, velocity, and acceleration. Li, et al. [63] put forward a model for the dynamic risk potential field of driving in a CAV environment, taking into account vehicles' acceleration and steering angle. Additionally, they developed a car-following model that considers the weight of surrounding vehicles. However, the studies using the risk field model assume that the AV's strategy for driving is to navigate through the field by following its valley [64] and driving efficiency is ignored, which may lead to overly dithering and inefficient driving decisions.

The probabilistic approaches rely on a probabilistic characterization of the temporal and spatial correlations between vehicles, while also factoring in the uncertainties associated with input data in safety risk evaluation. Li, et al. [50] proposed a probabilistic risk assessment method using safety metrics based on position uncertainty and distance to assess the driving risk. Chen, et al. [58] used a Kalman filter model to predict the probability of collision in the next monitoring interval. Naranjo, et al. [59] used a fuzzy logic approach to mimic human behavior and reactions during overtaking maneuvers. The probabilistic approaches solely rely on the expertise of specialists to produce rule-based actions and are incapable of making accurate decisions in the presence of environmental disturbances [50]. Mokhtari and Wagner [65] developed a probabilistic risk-based decision-making framework that included the probability and an estimate of an undesirable outcome, taking the number of pedestrians and the value of a loss of life into account. Noh [17] utilized a threat measure called TTE and Bayesian networks to assess the possibility of a collision of relevant vehicles. However, the probabilistic approach focuses only on assessing the likelihood of accidents and ignores the impact of the consequences of an accident on driving decision-making.

Therefore, this study utilizes the risk field approach to assess AVs' driving risks considering risks from all surrounding vehicles, and their weights are incorporated into the risk

><

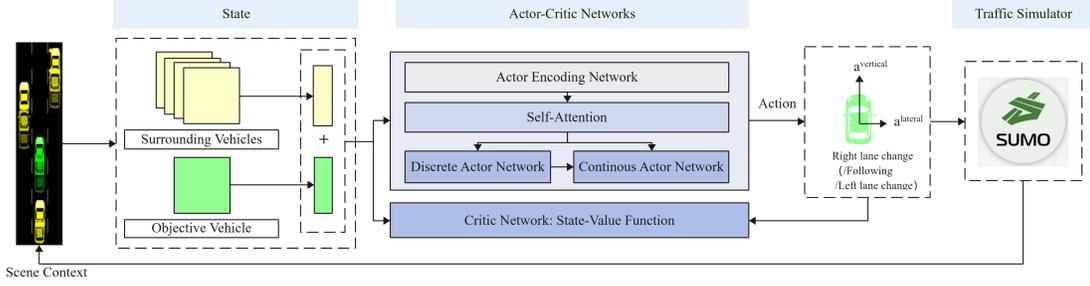

**Figure 1.** The architecture of the AT-HPPO decision-making model

assessment to account for the consequences of the potential accidents. Both driving risk and efficiency are considered in the AV's driving strategies.

## III. METHODOLOGY

In this section, we first introduce our overall approach to learn a driving model and the hybrid action space designed for driving behaviors. In addition, attention mechanisms for improving the performance of the algorithm are considered. Then, based on the driving risk field, we present the ADR indicator for evaluating the threats of surrounding traffic participants and take vehicle weights into account. Finally, we dive into driving decision-making strategy, considering the anticipated driving risk combined with the DRL algorithm.

### A. Hierarchical actor-critic-based DRL architecture

A decision strategy based on DRL allows an agent to operate in a stochastic environment, selecting a sequence of actions across multiple time steps while utilizing feedback (or rewards) to maximize the cumulative reward. Typically, this process is represented as a Markov decision process (MDP).

To enhance the stability and efficiency of training, the actor-critic framework [66] utilizes two neural networks. This structure separates the process of action evaluation and selection into distinct deep neural networks: an actor network and a critic network. Fig. 1 shows the hierarchical actor-critic framework used for this study.

Unlike the single actor of PPO, the actor network used in HPPO aims to make hybrid driving decisions, decomposed into two parts: discrete decisions at the upper level and continuous decisions at the lower level, with links between them. Car-following and lane-changing are the two common driving behaviors. The discrete decisions of AVs used in this study are whether to make a lane change or keep following, and the continuous decisions are how to make a lane change or conduct a car-following action.

The discrete decisions of AVs include continuing to follow, making a right lane change, and making a left lane change, which allows for more realistic driving in three dimensions. Then, the continuous planning for each discrete decision is learned. We define the following parameterized action space: the discrete actions are selected from a finite set: $A_d = \{a_{leftchange}, a_{following}, a_{rightchange}\}$, and each action $a_d \in A_d$ contains two continuous-valued actions, $a_{a_d}^{vertical} \in R_{a_d}^{vertical}$ and $a_{a_d}^{lateral} \in R_{a_d}^{lateral}$, which are the vertical accel-

eration and the lateral acceleration corresponding to $a_d$. The whole hybrid action space is defined as:

$$A = \bigcup_{a_d \in A_d} \{(a_d, a_{a_d}^{vertical}, a_{a_d}^{lateral}) \mid a_{a_d}^{vertical} \in R_{a_d}^{vertical}, a_{a_d}^{lateral} \in R_{a_d}^{lateral}\} \quad (1)$$

The system aims to learn to choose from the three discrete action decisions and apply the appropriate parameters for that chosen action. The single critic network is a state-value function, working as an estimator of the actor to evaluate how well the actor has performed.

### B. HPPO algorithm

We developed an HPPO algorithm to solve the problem of how to select a pair of actions, including a discrete action and two continuous actions, from the hybrid action space, based on the research of Fan, et al. [67] and the PPO algorithm.

PPO is a policy optimization method that learns a stochastic policy $\pi_\theta$ related to the parameter set $\theta$ of an actor network by minimizing a clipped surrogate objective [68]. Similar to the PPO algorithm, in each iteration of training, HPPO runs according to policies $\pi_{\theta_d}$ and $\pi_{\theta_c}$ in the environment, for $T$ time steps, and updates these two policies using the collected samples. The discrete policy $\pi_{\theta_d}$ and the continuous policy $\pi_{\theta_c}$ are updated separately based on minimizing their respective clipped surrogate objectives. The direction of lateral acceleration is constrained based on the selected discrete action during continuous action selection.

In the training process, we use the generalized advantage estimator (GAE) [69] method to estimate the advantage function $\hat{A}_t(a_t, s_t)$. Referring to recommended values [69], we set $\gamma$ and $\lambda$ to 0.95. The long short-term memory (LSTM) neural network is used in both actor network and critic network, as it gives well performance in processing time-series data.

For each layer of the algorithm architecture, the input higher-level task is broken down into subtasks and passed on to the next layer. Then, these two levels of reasoning are combined in one network with a hierarchical structure, as shown in Fig. 1, to retain the advantages of end-to-end training.

### C. Attention mechanism for DRL

When making driving decisions, human drivers can consider a sequence of past observations, assigning significance based on the time and location of each observation. An attention mechanism is integrated into the HPPO algorithm to achieve this capability, as illustrated in Fig. 2. Given our utili-zation of time series data, temporal attention is employed in this study to determine which frames hold the most significance in past observations. Spatial attention is not explicitly addressed, as it is typically employed with image data. However, the influence of location on importance is indirectly accounted for, as the time series data utilized contains relative location information.

We incorporate a temporal self-attention mechanism [70] to weigh the significance of past data in determining the current driving policy. To improve the selection of actions, we have integrated an attention module into the actor network, with weights defined as:

$$\text{Attention}(Q, K, V) = \text{softmax}\left(\frac{QK^T}{\sqrt{d_k}}\right)V \tag{2}$$

where the three matrices, Query (Q), Value (V), and Key (K), are learned linear transformations of the input $[h_t, h_{t+1}, ..., h_T]$ to embed contextual information. It computes a weights matrix based on the relevance of different parts in $Q$ and $K$, and subsequently adjusts the weights of parts in $V$ accordingly. $T$ of $h_T$ is the total time of our input trajectory information.

### D. Anticipated driving risk considering vehicles' weights

Regarding the risk assessment module, it is necessary to evaluate the threats posed by all surrounding traffic participants. This study uses field theory and establishes a dynamic driving risk field model considering vehicles' weights, based on Li, et al. [63]. They refer to the dynamic risk field as a kinetic field, which is determined by the moving objects on roads. However, their work only establishes a car-following model based on the potential driving risk field. We extended their work to both car-following and lane-changing situations and proposed a new indicator based on potential field theory.

Li, et al. [63] defined attributes of vehicles in the vehicle's potential field are the vehicle's speed and actual mass. Based on the data in the paper [71], it can be seen that the willing-ness and risk of following light and heavy vehicles do not increase in equal proportion with vehicle weight, so referring to the literature, we add $T_A$ to correct. The vehicle's kinetic field strength $E_k^{AB}$ from SV $A$ to OV $B$ can be expressed as:

$$E_k^{AB} = T_A \cdot m_A \cdot (1.566 \times 10^{-14} \cdot v_A^{6.687} + 0.3345)$$
$$\cdot \lambda \frac{e^{-\beta_1 a_A \cos\alpha}}{|k'_{AB}|} \cdot \frac{k'_{AB}}{|k'_{AB}|} \tag{3}$$

$$\left| k'_{AB} \right| = \sqrt{\left[ (x_B - x_A) \frac{\tau}{e^{av_A}} \right]^2 + \left[ (y_B - y_A)\tau \right]^2} \tag{4}$$

The definitions of the variables are all listed in TABLE III in Appendix A. $T_A$, $\beta_1$, $\beta_2$, $\lambda$, and $\tau$ are the undetermined coefficients.

Leveraging the potential field framework, we optimize and extend the formulation of field force for the integration of risk field force calculations for lane-changing and following states. Besides the motion state of the surrounding vehicle (SV), it is evident that the magnitude of the field force acting on the object vehicle (OV) varies not only with the velocity and acceleration gap between the OV and the SV but also with the directions of motion of both vehicles. The field force $F_{AB}$, based on the vehicle's kinetic field strength from SV $A$ to OV $B$ can be calculated by (5).

$$F_{AB} = T_B \cdot m_B \cdot (1.566 \times 10^{-14} \cdot v_A^{6.687} + 0.3345) \cdot$$
$$e^{-(\beta_2(v_B \cos\theta - v_A) + \beta_3(a_B \cos\gamma - a_A)) \cdot \cos\alpha} \cdot E_k^{AB} \tag{5}$$

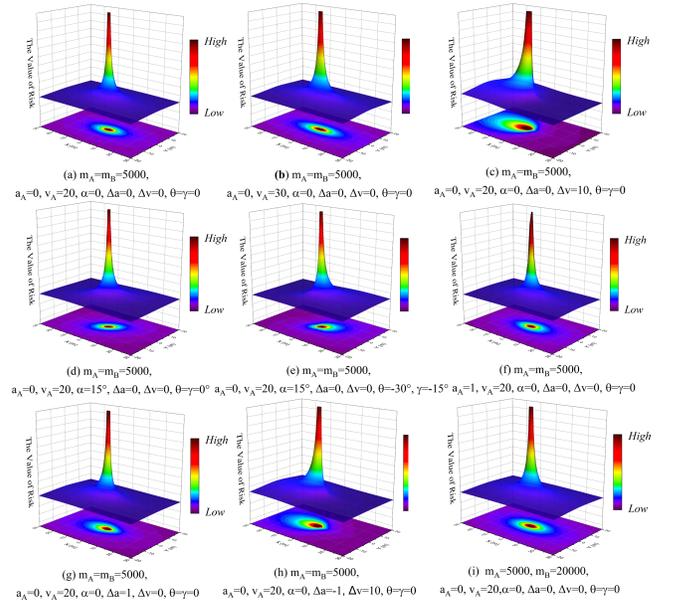

(a) $m_A=m_B=5000$,
$a_A=0, v_A=20, \alpha=0, \Delta a=0, \Delta v=0, \theta=0$

(b) $m_A=m_B=5000$,
$a_A=0, v_A=30, \alpha=0, \Delta a=0, \Delta v=0, \theta=0$

(c) $m_A=m_B=5000$,
$a_A=0, v_A=20, \alpha=0, \Delta a=0, \Delta v=10, \theta=0$

(d) $m_A=m_B=5000$,
$a_A=0, v_A=20, \alpha=15°, \Delta a=0, \Delta v=0, \theta=0°$

(e) $m_A=m_B=5000$,
$a_A=0, v_A=20, \alpha=15°, \Delta a=0, \Delta v=0, \theta=30°, \gamma=15°$

(f) $m_A=m_B=5000$,
$a_A=1, v_A=20, \alpha=0, \Delta a=0, \Delta v=0, \theta=0$

(g) $m_A=m_B=5000$,
$a_A=0, v_A=20, \alpha=0, \Delta a=1, \Delta v=0, \theta=0$

(h) $m_A=m_B=5000$,
$a_A=0, v_A=20, \alpha=0, \Delta a=1, \Delta v=10, \theta=0$

(i) $m_A=5000, m_B=20000$,
$a_A=0, v_A=20, \alpha=0, \Delta a=0, \Delta v=0, \theta=0$

**Figure 3.** Distribution and surface projection of risk field force corresponding to various conditions. The lane direction aligns with the X-axis direction. The X-value represents the longitudinal distance of the OV from the vehicle in the lane, while the Y-value represents the lateral distance. SV is set at (0, 0) and the risk field force reaches its maximum at this point and approaches infinity. $\Delta a$ and $\Delta v$ are the acceleration and velocity gap between OV and SV. The definitions of $m_A, m_B, a_A, v_A, \alpha, \theta$ and $\gamma$ are consistent with those provided in TABLE III in Appendix A.

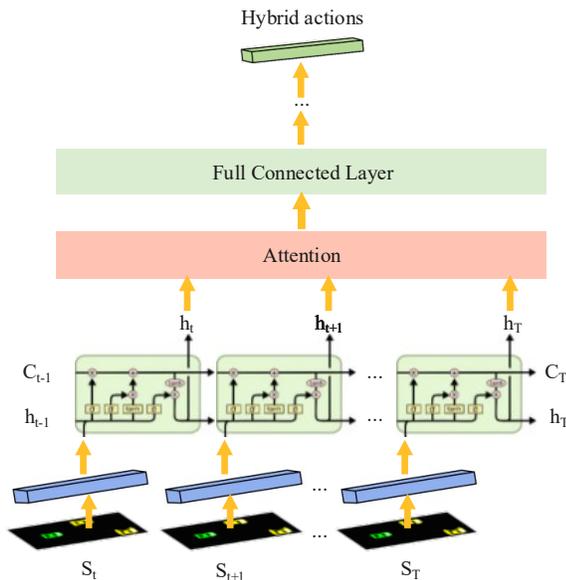

**Figure 2.** The architecture of actor network in the AT-HPPO.

Fig. 3 illustrates the distribution and surface projection of risk field force corresponding to various conditions. Initially, we observed a consistent impact of distance on the risk field force. The risk field force diminishes as the relative distance between vehicles increases. Once the relative distance exceeds a certain critical value, the risk field force becomes negligible. However, as the distance approaches the SV's position, the risk field force increases indefinitely. Undoubtedly, the extent to which other factors affect the risk field force varies, and the impact of these factors is analyzed as follows.

1) When comparing Fig. 3 (a) and Fig. 3 (b), as well as Fig. 3 (a) and Fig. 3 (i), where the SV moves at a constant speed along the x-axis, it becomes evident that the velocity and weight of the SV amplify the risk field force on the OV. The enhancement along the X-axis is more pronounced than the effect on the Y-axis, which remains constant.

2) In Fig. 3 (a), Fig. 3 (g), and Fig. 3 (c), positive values of both relative velocity and relative acceleration result in a displacement of the risk force distribution towards the rear of the SV, rendering the rear of the SV as the primary high-risk region.

3) The comparison between Fig. 3 (a) and Fig. 3 (d) reveals that the distribution of risk forces rotates following the direction of travel, particularly when the direction of SV changes and the velocity directions of SV and OV align.

4) The discrepancy between Fig. 3 (d) and Fig. 3 (e) arises from their different motion directions, leading to a redistribution of risk field force. The disparate motion directions lead to the OV having lower velocity and acceleration in the direction of SV motion compared to the SV, thereby increasing the risk when the OV precedes the SV's motion direction.

5) Upon comparing Fig. 3 (f) and Fig. 3 (a), it is evident that the risk field force distribution of the latter is slightly skewed towards the front, indicating that the influence of the OV ahead of the SV is slightly greater than that on the rear during acceleration of SV.

6) Upon comparison between Fig. 3 (h) and Fig. 3 (c), it is evident that a relative acceleration of -1 can significantly diminish the magnitude of the risk for the OV traveling behind the SV when the relative velocity is greater than zero.

As OVs are influenced by many surrounding vehicles while traveling on the road, we sum up these risks and propose an indicator of anticipated driving risk (ADR):

$$ADR = \sum_{i=1}^{N} \left| F_{S_i B}^i \right| \qquad (6)$$

where $N$ is the number of SVs and $i$ is the index of SVs. $F_{S_i B}^i$ is defined in (5). For some values of the undetermined parameters, we refer to [63].

### E. Risk anticipation-based decision making

*State space and action space*

The environmental data describes the environmental states that AVs need to obtain to form a strategy while the vehicle is driving. The main environmental data when driving includes (1) information about SVs' states, such as speed gap, position gap to OV, direction, etc. and (2) information about the OV's state, such as speed, direction, acceleration, etc.

To enhance the learning performance when dealing with multiple inputs and thereby make optimal decisions during driving based on the relationship between the OV and SVs, we simplify the inputs to reduce the complexity without losing too much information. Relative distance (longitudinal and lateral), velocity (longitudinal and lateral), and yaw angle are considered comprehensively in decision learning. The relationship between the states of the OV and one of the SVs can be written as:

$$SV_i = \left\{ \Delta d_i^{ver}, \Delta d_i^{la}, \Delta v_i^{ver}, \Delta v_i^{la}, yaw_i, x_i \right\}, x_i \in [0,1] \quad (7)$$

whose variables are explained in TABLE IV in Appendix B. The relative speed is added to enable the network to predict the future driving states of SVs to learn basic driving skills such as obstacle avoidance. In this study, sound information regarding SVs, such as the horn, engine sound, etc., is ignored.

The main aim is to explore the driving strategies of the OV. Therefore, the focus is on the decision to drive rather than on the dynamics of the driving model. The OV's motion is updated based on vertical acceleration and lateral acceleration using a simple differential process. The properties of the OV's attributes are expressed in (8) and the complete state space $S$ can be described as in (9), integrating the states of the OV and SVs. Information from six (N=6) SVs, including both the following and leading vehicles in the current lane and adjacent lanes, is utilized [72, 73]. If necessary, we can accommodate more vehicles by adjusting the state space.

$$OV = \{v_{ver}, v_{la}, a, pos_{ver}, yaw, l_l, l_r\}, l_l, l_r \in [0,1] \quad (8)$$
$$S = [SV_1, SV_2, \ldots, SV_N, OV] \qquad (9)$$

For the three discrete decisions of continuing to follow, left lane changing and right lane changing, it is assumed that they follow a categorical distribution. For the continuous driving decisions regarding the target acceleration, it is assumed that they follow a clipped normal distribution [68, 74], with each dimension independent of the other. The mean and variance of the normal distribution are the outputs from the neural network of the continuous actor.

*Reward function*

In RL, the formalization of the goal of the agent is embodied as a reward signal. At each step, the reward signal is real-valued, $R \in \mathbb{R}$. Multi-target issues are very common when designing reward functions [75, 76]. Considering that driving should be a balance between efficiency and safety, and taking into account the smoothness of driving and traffic rules, the reward function is designed as follows:

$$
\begin{aligned}
R &= \omega_1 r^{risk} + \omega_2 r^{vertical} + \omega_3 r^{position} \\
&\quad + \omega_4 r^{limit} + \omega_5 r^{collision} \\
&= -\omega_1 \cdot ADR \cdot v_{ver} - \omega_3 \cdot pos_{ver} \\
&\quad - \omega_4 \cdot k \cdot (v_{ver} - v_{limit}) - \omega_5 \cdot r^{collision}
\end{aligned}
\quad (10)
$$

$r^{risk}$ encourages AV driving to minimize the anticipated driving risk. $r^{vertical}$ denotes the reward for AVs is to follow the planned and established route. $r^{position}$ guides vehicles to drive along the center line of the road when following, and prevents them from remaining in the center of the lane for an extended period when changing lanes. $r^{limit}$ and $r^{collision}$ are utilized to penalize vehicles for speeding and collisions, respectively. $k$ is a 0-1 variable that equals 1 when the OV exceeds the speed limit. We utilized max-min normalization for the first three terms to facilitate comparison of their significance in decision-making. $\omega_1, \omega_2, \omega_3, \omega_4, \omega_5$ denote the weight of each item, which are equal to 1, 2, 0.5, 100, and 100, respectively. The two penalty items ($r^{limit}$ and $r^{collision}$) are binary variables and are assigned significant weights to prevent the violations [77]. When the vehicle can comply with the speeding rules and avoid collisions, the two penalty items are equal to 0. By comparing the weights of the first three items, the weight of ADR which integrates the weights of surrounding vehicles equals 1, evidencing the importance of considering the weights of surrounding vehicles.

*Training details*

As explained, the HPPO algorithm is introduced to help find an optimal driving strategy with risk anticipation. The actor network is responsible for generating strategies and the critic network for evaluating them. The details of the networks are shown in TABLE V in Appendix C. ReLu is a non-linear activation function. To obtain better computational performance, the size of fully connected layers is chosen to be the integer power of 2.

Some tricks are used to improve the performance of the HPPO algorithm, as follows:

---

**Algorithm 1**: Hierarchical proximal policy optimization (HPPO) (training).

---

for $iteration \in \{1, \ldots, M\}$ do

  for $t \in \{1, \ldots, T\}$ do

    Run policy $\pi_{\theta_d}$ and $\pi_{\theta_c}$ for $T$ time steps, collecting $\{s_t, a_t^d, a_t^c, r_t, d_t\}$ for $t \in \{1, \ldots, T\}$

    Estimate advantages: $\hat{A}_t = \sum_{l=1}^{T-1}(\gamma\lambda)^l \delta_{t+l}$

    where $\delta_t = r_t + \gamma V_\phi(s_{t+1}) - V_\phi(s_t)$

    Estimate return: $\hat{R}_t = \hat{A}_t + V_\phi(s_t)$

    Store partial trajectory information

  end for

  $\pi_{old_d} \leftarrow \pi_{\theta_d}, \pi_{old_c} \leftarrow \pi_{\theta_c}$

  for $b \in \{1, \ldots, B\}$ do

    $J_{PPO2}^d(\theta) =$

    $\sum_{(s_t, a_t^d)} \min \left( \frac{\pi_{\theta_d}(a_t^d|s_t)}{\pi_{old_d}(a_t^d|s_t)} \hat{A}_t, clip\left( \frac{\pi_{\theta_d}(a_t^d|s_t)}{\pi_{old_d}(a_t^d|s_t)}, 1-\varepsilon, 1+\varepsilon \right) \hat{A}_t \right)$

    $J_{PPO2}^c(\theta) =$

    $\sum_{(s_t, a_t^c)} \min \left( \frac{\pi_{\theta_c}(a_t^c|s_t)}{\pi_{old_c}(a_t^c|s_t)} \hat{A}_t, clip\left( \frac{\pi_{\theta_c}(a_t^c|s_t)}{\pi_{old_c}(a_t^c|s_t)}, 1-\varepsilon, 1+\varepsilon \right) \hat{A}_t \right)$

    $L_{BL}(\phi) = \max \left( \sum_{t=1}^{T} \left( \hat{R}_t - V_\phi(s_t) \right), \sum_{t=1}^{T} \left( \hat{R}_t - V_\phi^{clip}(s_t) \right)^2 \right)$

    $L_{total} = J_{PPO2}^d(\theta) + J_{PPO2}^c(\theta) + 0.5 * L_{BL}(\phi)$

    $-0.01 * \left( Entropy(\pi_{\theta_d}) + Entropy(\pi_{\theta_c}) \right)$

    update $\theta$ by gradient descent optimization w.r.t $L_{total}$

    update $\phi$ by gradient descent optimization w.r.t $L_{total}$

  end for

end for

---

1) *Learning rate decay*: Prior research [78] has shown that learning rate decay can somehow enhance the smoothness of the later stages of training, and substantially improve the training results. Thus, we use a linear decay of the learning rate in Adam, with the learning rate decreasing linearly, from an initial 3e-4 to 0, with the number of training steps.

2) *Gradient clipping*: The gradient is clipped during backpropagation to avoid the explosion of the gradients. A predefined coefficient is set to 0.1 which sets the maximum of the standard deviation after clipping.

The corresponding pseudo-code for our method is shown in Algorithm 1. The variables in Algorithm 1 are explained in TABLE VI in Appendix D.

The training at each step of the simulation collects vehicle data and lane data from the environment, as well as the current state of the AV through the neural network, to generate a Gaussian distribution of actions, which is used to generate specific actions that feed into the simulation environment. Each simulation is reset when the maximum time range is reached or the stopping condition is met. Once enough training data has been collected, a sample is taken from the data for training. The value network is updated via the Bellman equation. The mini-batch size is 4. For calculating the value and advantage, the discount factor $\gamma$ is 0.99. The $\lambda$ used in $TD(\lambda)$ and $GAE(\lambda)$ is 0.95.

## IV. EXPERIMENTS

This section describes the simulation experiment setting when training by SUMO [27] simulator, evaluation metrics, and comparison algorithms.

*A. Experiment settings*

The risk-anticipation-based AV driving strategy considering vehicles' weights is especially important in contexts with a high proportion of heavy vehicles. Therefore, an example of a three-lane highway is applied to illustrate the effect of our proposed approach. The length of the highway is configured to be 2.8 km, and the proportion of heavy vehicles is set to approximately 25%, based on the proportion in the HighD dataset [28].

The traditional vehicles are controlled by the internal model of SUMO, which is designed to mimic real human driving behaviors. Krauss and LC2013 [79], respectively, are the default car-following and lane-changing models used in SUMO to control human-driven vehicles. To accurately simulate actual traffic flow conditions, we made several adjustments to the LC2013 model parameters [79, 80]. Specifically, we increase the eagerness for performing strategic lane changing and lane changing to gain speed. We mitigate the risk of traffic congestion caused by excessively cautious lane change behavior by utilizing the selected combination of lane change model parameters.

Traffic flows are introduced at various points, maintaining Poisson distribution ($\lambda$=1.2, veh/s) following the HighD dataset for each trial epoch during training. We randomly initialize the speeds of the surrounding vehicles to enhance the road complexity and set the initial speed of all vehicles to 20 m/s. The simulation warm-up period is set to 115 s to stabilize the road conditions and prevent catastrophic congestion resulting from the low initial speeds of most vehicles. Additionally, the speed limit of the road is set to 120 km/h.

Finally, we validate the algorithms in the simulation environments under varying traffic flow densities and using the highD dataset [28], respectively.

### B. Evaluation metrics

Driving efficiency is evaluated based on the calculation of average speed. The safety-related assessment includes two parts: evaluation of the risk likelihood and the consequences of potential collisions.

To evaluate the likelihood of potential risks, the number of conflicts is utilized, with surrogate measures of TTC (time to collision) [81], DRAC (deceleration rate to avoid a crash) [82], and PET (post encroachment time) [83]. Their thresholds are set to $3.0s$, $3.0m/s^2$ and $2.0s$ respectively [84]. Their calculation formulas can be found in these papers [81-83].

To estimate the consequences of potential collisions, we create a new indicator, potential collision energy in conflicts (PCEC), based on potential collision energy (PCE) [85], which is used to account for the effect of momentum. A modified PCE was proposed by Wang, et al. [71]:

$$PCE_i(t) = \begin{cases} \frac{1}{2}\alpha_i\alpha_f\left[m_f v_f^2(t) - m_i v_i^2(t)\right], \; m_f v_f^2(t) - m_l v_l^2(t) > 0 \\ \frac{1}{2}\alpha_i\alpha_f m_f v_f^2(t), \; m_f v_f^2(t) - m_l v_l^2(t) \leqslant 0 \end{cases} \quad (11)$$

where $m_f$ and $m_l$ denote the masses of the following vehicle and leading vehicle respectively. $v_f$ and $v_l$ denote the velocities of the following and leading vehicles respectively. $\alpha_f$ and $\alpha_l$ denote the attribute functions of the following and leading vehicles, respectively. For instance, vehicle dimensions, maneuverability, and acceleration and deceleration performance are considered. Referring to [71], $\alpha_l$ and $\alpha_f$ are set to 1. Since distance and angle information is not accounted for in (11), we propose the PCEC to evaluate the consequences of potential collisions in one epoch:

$$PCEC = \sum_t^T \sum_i^N y_{it} \cdot PCE_i(t) \quad (12)$$

$y_{it}$ is a 0-1 variable that equals 1 when these two vehicles are in conflict.

### C. Comparison Baselines

For the baseline, we selected HPPO without attention (denoted as HPPO), HPPO using TTC instead of ADR (denoted as HPPO-T), DDPG, PPO, Soft Actor-Critic (SAC) and rule-based model (IDM+LC2013) to sufficiently investigate the advantages of HPPO, ADR and attention mechanism. In particular, except for our proposed HPPO with attention (AT-HPPO), none of the other methods include an attention mechanism.

## V. RESULTS AND DISCUSSION

### A. Evaluation of the Learning Capability

The training performance of the agent is evaluated by evaluating the episodic reward in this section. Fig. 4 illustrates the average learning curves of five approaches. During the convergence phase, fluctuations in cumulative rewards were primarily attributed to noise in the driving environment. It is observed that the proposed AT-HPPO and HPPO outperformed comparison baselines in both sample efficiency and robustness after quick convergence. Additionally, the attention mechanism accelerates convergence and helps the algorithm achieve higher reentry rates and enhance robustness.

### B. Performance Evaluation in Simulation

To further evaluate the effectiveness of the proposed AT-HPPO method, we conduct a comparison with other baseline methods across different road conditions with three traffic densities. The HPPO-T method utilizes TTC instead of ADR in the reward function, while all other methods employ the same reward function. The average densities of sparse, medium, and dense traffic are 11.52 (veh/km), 25.57 (veh/km), and 32.91 (veh/km) respectively. All experimental results were conducted through six independent trials, each using a different random seed, for statistical evidence. The average values from the test simulation are shown in TABLE I.

Firstly, it is observed that through training, all methods successfully avoid collisions and adhere to speed limits in various scenarios. In dense and medium densities, the HPPO algorithm consistently exhibits superior performance in terms of average speed, number of lane changes, PCEC and ADR. The reduced number of conflicts, as well as the lower PCEC and ADR metrics, demonstrate that the HPPO algorithm can substantially diminish the probability and severity of accidents.

At sparse traffic, HPPO, despite enabling more frequent lane changes, did not result in an improvement in driving efficiency, as indicated by its average driving speed of 27.29, which is lower than the average speed of 31.58 obtained by the IDM+LC2013 method. The AT-HPPO achieved the same number of lane changes (3.17) and demonstrated an improved average speed of 31.98 m/s compared to the HPPO. This suggests that attentional mechanisms assist in enhancing the accuracy of decision-making processes.

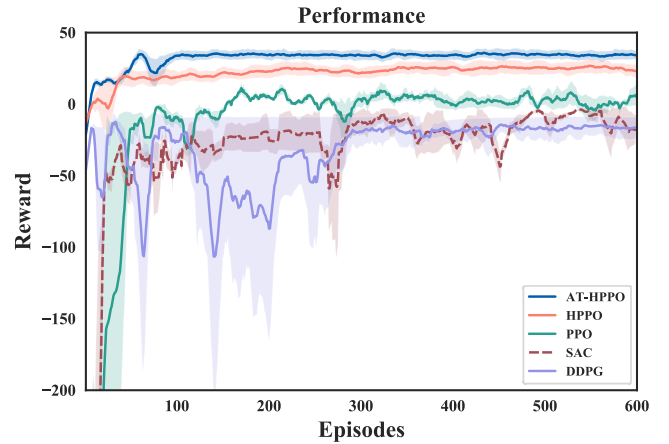

**Figure 4.** Learning curves of five approaches.

TABLE I
VERAGE VALUES FROM TEST SIMULATION.

| | IDM+LC2013 | DDPG | PPO | SAC | HPPO-T | HPPO | AT-HPPO |
|---|---|---|---|---|---|---|---|
| **Sparse Traffic (density of 11.52 veh/km)** | | | | | | | |
| Average speed (m/s) | 31.58 | 20.38 | 20.73 | 20.34 | 22.77 | 27.79 | **31.98** |
| Number of lane changes | 2 | 4 | 3 | 3 | 2 | **3.17** | **3.17** |
| Collision | **0** | **0** | **0** | **0** | **0** | 0 | 0 |
| Number of conflicts | **0** | 1.75 | 3.83 | 2.17 | 3.33 | 0 | 0 |
| Number of heavy vehicles in conflicts | **0** | 0.75 | **0** | **0** | 1.33 | 0 | 0 |
| Number of light vehicles in conflicts | **0** | 1 | 3.83 | 2.17 | 2 | **0** | **0** |
| PCEC (kJ, 10e2) | **0** | 248.05 | 176.03 | 87.76 | 184.40 | **0** | **0** |
| ADR (10e2) | 0.62 | 1.6 | 1.74 | 0.97 | 1.46 | 0.31 | 0.34 |
| **Medium Traffic (density of 25.57 veh/km)** | | | | | | | |
| Average speed (m/s) | 26.53 | 19.30 | 18.41 | 14.59 | 20.63 | 26.72 | **28.55** |
| Number of lane changes | 2 | 4 | 2.83 | 3.5 | 2.33 | **3.5** | 3.17 |
| Collision | **0** | **0** | **0** | **0** | **0** | **0** | **0** |
| Number of conflicts | 1.5 | 3 | 4.33 | 4.17 | 2.17 | 1.17 | **1.03** |
| Number of heavy vehicles in conflicts | **0** | **0** | **0** | **0** | 1 | **0** | **0** |
| Number of light vehicles in conflicts | 1.5 | 3 | 4.33 | 4.17 | 1.17 | 1.17 | **1.03** |
| PCEC (kJ, 10e2) | 95.06 | 132.21 | 316.34 | 103.17 | 146.35 | 89.78 | **80.95** |
| ADR (10e2) | 0.67 | 0.65 | 2.79 | 1.34 | 2.51 | 0.57 | **0.55** |
| **Dense Traffic (density of 32.91 veh/km)** | | | | | | | |
| Average speed (m/s) | 21.20 | 15.29 | 18.66 | 15.49 | 19.92 | 22.44 | **26.12** |
| Number of lane changes | 2 | 3.17 | 2.83 | 3.17 | 2 | **2.33** | 2.17 |
| Collision | 0 | 0 | 0 | 0 | 0 | 0 | **0** |
| Number of conflicts | 1 | 7.5 | 5.83 | 2.17 | 4.34 | 0.66 | **0.33** |
| Number of heavy vehicles in conflicts | 0.67 | **0** | 1.5 | 0.67 | 2.67 | **0.33** | **0.33** |
| Number of light vehicles in conflicts | 0.33 | 7.5 | 4.33 | 1.5 | 1.67 | 0.33 | **0** |
| PCEC (kJ, 10e2) | 92.51 | 352.46 | 353.83 | 115.02 | 397.03 | 46.58 | **21.99** |
| ADR (10e2) | 0.62 | 0.73 | 3.04 | 1.52 | 2.19 | 0.92 | **0.57** |

HPPO significantly outperforms HPPO-T in all three scenarios. Analyzing the dense traffic scenario as an example, comparing HPPO-T and HPPO, a significant decrease in the number of conflicts is observed, from 4.33 to 0.67, among which the number of heavy vehicles involved in conflicts decreases from 2.67 to 0.33 and the number of light vehicles decreases from 1.67 to 0.33. The number of lane changes increase, from 2 to 2.33. The driving efficiency is increased at the same time, with the average speed of the AV increasing from 19.92m/s to 22.44m/s. The consequences of potential collisions decrease substantially, as seen by the PCEC dropping from 39,703kJ to 4,658kJ. The significant improvements in all of the performance measures evidence that the inclusion of ADR in the reward function can yield better AV driving strategies, leading to significant reductions in both the likelihood and consequences of potential collisions and an improvement in AV driving efficiency at the same time.

Specifically, AT-HPPO emerges as the top-performing algorithm, achieving a greater number of lane changes and demonstrating the efficacy of these changes through optimization of both the highest average speed and the lower number of conflicts. Based on the evaluation metrics, we can conclude that the hierarchical reinforcement learning approach employed by HPPO, combined with attention, can decrease the likelihood of accidents (by reducing potential conflicts and involving fewer heavy vehicles in conflicts) and mitigate the severity of conflicts (by decreasing PCEC), while simultaneously enhancing driving efficiency, as evidenced by an increased average speed.

### C. Performance Evaluation in Dataset

RL policies are prone to unsafety when confronted with unknown or unseen traffic scenarios [86]. To demonstrate the robustness of the proposed AT-HPPO method and evaluate its performance in real-world scenarios, the HighD dataset [28], featuring some similarities in simulation scenarios, was utilized. In each experiment, a car was randomly selected and controlled using either the HPPO or AT-HPPO algorithm. The experimental data, obtained by repeating this process thirty times, are presented in TABLE II.

The data reveals that the metrics of AT-HPPO outperform those of human drivers in similar but untrained new environments, indicating the robustness and generalization capabilities of our proposed algorithm, which excels at making appropriate decisions tailored to the environment. Comparing HPPO with human drivers, the speed increases from 29.69 m/s to 30.06 m/s, the PCEC decreases from 53,483kJ to 51,657kJ, and the ADR decreases from 8.15 to 7.71. However, in comparison to AT-HPPO, although HPPO increases the number of lane changes from 7 to 9, the efficiency and safety metrics are inferior, suggesting that the integration of attention mechanisms with HPPO facilitates intelligent decision-making for safe and efficient driving behaviors. We present some of the experimental tracks, as depicted in Fig. 5, to facilitate a more comprehensive comparison. We select a typical scenario, specifically, driving behavior where the speed of the vehicle behind OV exceeds that of OV.

Upon comparing Fig. 5. (b)-(d), it is evident that both the human driver and the AT-HPPO can execute lane change maneuvers swiftly and accurately to avoid being rear-ended. However, the HPPO exhibits prolonged lane change times and prolonged proximity to the lane line, posing potential safety hazards. Upon comparing the track information of the human driver and AT-HPPO, it becomes evident that AT-HPPO exhibits greater stability in both lateral displacement and velocity control processes.

TABLE II
AVERAGE VALUES FROM TEST IN DATASET.

|  | Drivers | HPPO | AT-HPPO |
|---|---|---|---|
| Average speed (m/s) | 29.69 | 30.06 | 30.57 |
| Number of lane changes (summation) | 5 | 9 | 7 |
| Number of conflicts | 4 | 3.84 | 3.55 |
| Number of heavy vehicles in conflicts | 1.47 | 1.26 | 1.35 |
| Number of light vehicles in conflicts | 2.53 | 2.58 | 2.2 |
| PCEC (kJ, 10e2) | 534.83 | 516.57 | 489.18 |
| ADR (10e2) | 8.15 | 7.71 | 5.57 |

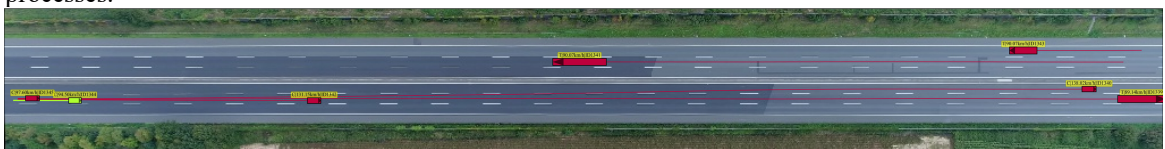

(a) Initial State. The OV is numbered 1344 and has an initial speed of 94.50 km/h. It is followed by vehicle 1345, traveling at 97.60 km/h.

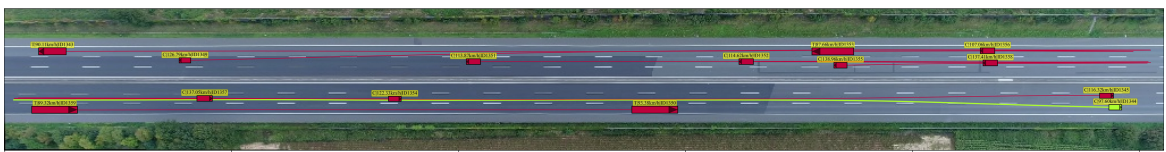

(b) Track of Human Driver

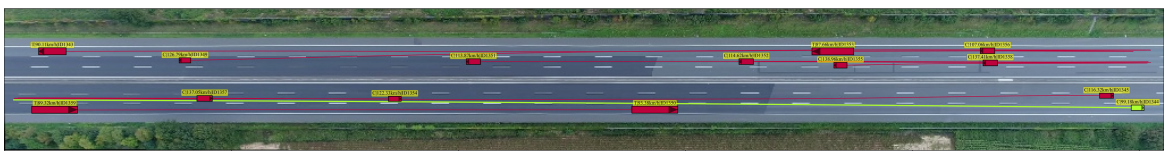

(c) Track of HPPO

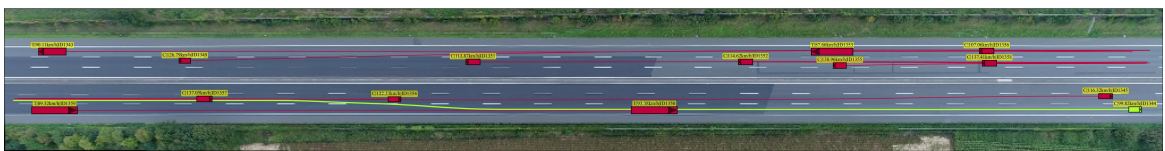

(d) Track of AT-HPPO

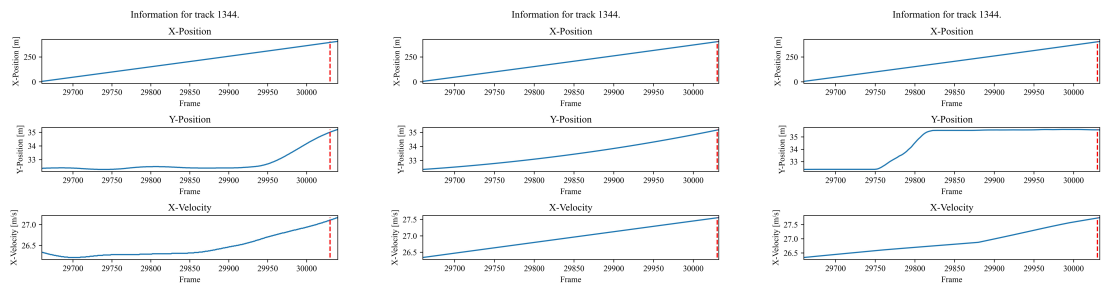

(e) Information for tracks. (From left to right are the track information of the human driver, HPPO, AT-HPPO)

**Figure 5.** Driving tracks of different methods. The tracks and vehicles of OV are depicted in green, while SVs and their tracks are displayed in red.

## VI. Conclusions

This paper proposes a risk anticipation- and hierarchical DRL-based AV driving strategy, considering the weights of the surrounding vehicles of AVs. The main findings and conclusions can be summarized as follows:

1) The hierarchical DRL framework, featuring mixed action decision-making comprising discrete decisions on whether to continue following or changing lanes and continuous decisions on how to proceed with lane following or changing, enables the AV to effectively assess the environment and make appropriate decisions, thus achieving a balance between efficiency and safety and enhancing the robustness and generalization capabilities.

2) The proposed ADR considers the weight of all surrounding vehicles, and its integration into the driving strategy aids in diminishing both the total number of conflicts and those involving heavy vehicles. This reduction can mitigate the likelihood and severity of accidents, thereby enhancing safety.

3) The proposed reward function in DRL, incorporating five components, namely anticipated risk, following the planned route, staying in lane, and penalties for overspeeding and collisions, enables the optimized AV driving strategy to perform well with low driving risk, high driving efficiency, high stickiness to the planned route, and a safe driving speed.

4) Integrating the attention mechanism with HPPO (AT-HPPO) can further enhance the robustness and generalization capabilities, leading to improved performance and higher rewards. Furthermore, it contributes to refining the AV driving strategy, thereby reducing the likelihood and consequences of potential conflicts and enhancing driving efficiency.

The experiment conducted in this study shows the feasibility and efficiency of our proposed AV strategy-making method. We expect that our method could contribute substantially to AV driving control, especially in the context of a high proportion of heavy vehicles or mixed traffic flow with a variety of vehicle types. In future work, we plan to enhance our methodology by combining with perception algorithms, exploring advanced algorithms for accurate estimation of vehicle weight, and exploring the implementation of safe HRL. We also plan to explore various scenarios, encompassing different speed limits, traffic zones, and more.

## Appendix

### A. Definition of Variables

TABLE III
DEFINITION OF VARIABLES

| Variable | Definition |
|---|---|
| $m_B$ | weight of OV $B$ |
| $v_B$ | velocity of OV $B$ |
| $m_A$ | weight of SV $A$ |
| $v_A$ | velocity of SV $A$ |
| $\theta$ | clockwise angle formed by the direction of the velocity of SV $A$ and OV $B$ |
| $\gamma$ | clockwise angle formed by the direction of acceleration of SV $A$ and OV $B$ |
| $\alpha$ | clockwise angle formed by the mass center of OV $B$ and SV $A$ with the lane |
| $a_A$ | acceleration of current motion state of SV $A$ |
| $x_A, y_A$ | horizontal and vertical coordinate values of mass center of $A$ on road |
| $x_B, y_B$ | horizontal and vertical coordinate values of mass center of $B$ on road |

### B. Description of State Space

TABLE IV
DESCRIPTION OF STATE SPACE

| Variable | Description |
|---|---|
| $i$ | index of SVs |
| $N$ | total number of SVs |
| $\Delta d_i^{ver}$ | relative vertical distance between OV and $SV_i$ |
| $\Delta d_i^{la}$ | relative lateral distance between OV and $SV_i$ |
| $\Delta v_i^{ver}$ | relative velocity between OV and $SV_i$ in vertical direction |
| $\Delta v_i^{la}$ | relative velocity between OV and $SV_i$ in lateral direction |
| $yaw_i$ | vehicle's (OV or $SV_i$'s) yaw angle |
| $x_i$ | whether $SV_i$ is in the perception range $(+50m, -50m)$ |
| $v_{ver}$ | velocity of OV in vertical direction |
| $v_{la}$ | velocity of OV in lateral direction |
| $a$ | acceleration of OV |
| $pos_{ver}$ | lateral position of OV in its current lane |
| $l_l$ | whether the OV is in the leftmost lane |
| $l_r$ | whether the OV is in the rightmost lane |

### C. Details of Actor and Critic Networks

TABLE V
DETAILS OF ACTOR AND CRITIC NETWORKS

| Layer | Hidden units | Activation |
|---|---|---|
| I | 64 | ReLu |
| II | 128 | ReLu |
| III(LSTM) | 64 | |
| IV | 32 | ReLu |
| Output I (continuous) | 2 | Linear |
| Output II (discrete) | 1 | Linear |
| Output III (critic) | 1 | Linear |

### D. Variables in Algorithm 1

><

## TABLE VI
### VARIABLES IN ALGORITHM 1

| Variable | Definition |
|---|---|
| $s_t$ | OV state at time $t$ |
| $a_t^d, a_t^c$ | discrete action and continuous action of OV at time $t$ |
| $r_t$ | reward at time $t$ |
| $d_t$ | 0-1 variable denoting whether a training epoch is over |
| $V_\phi(s_t)$ | output of state-value function with parameter $\phi$ |
| $B$ | number of iterative steps needed to update actor and critic networks given one batch of data |
| $\pi_{\theta_d}(a_t^d \| s_t)$ | probability of choosing $a_t^d$ under $s_t$ with parameter $\theta_d$ |
| $\pi_{old_d}(a_t^d \| s_t)$ | probability of choosing $a_t^d$ under $s_t$ with parameter $old_d$ |
| $\pi_{\theta_c}(a_t^c \| s_t)$ | probability of choosing $a_t^c$ under $s_t$ with parameter $\theta_c$ |
| $\pi_{old_c}(a_t^c \| s_t)$ | probability of choosing $a_t^c$ under $s_t$ with parameter $old_c$ |
| $\theta$ | parameter set of actor network including $\theta_d$ and $\theta_c$ |
| $\phi$ | parameter set of critic network |

## REFERENCES

[1] NHTSA, "Early Estimates of Motor Vehicle Traffic Fatalities For thee First Half (January-June) of 2023," National Highway Traffic Safety Administration, U.S. Department of Transportation, Washington, DC, U.S., 2023.

[2] N. Nadimi, H. Behbahani, and H. Shahbazi, "Calibration and validation of a new time-based surrogate safety measure using fuzzy inference system," *Journal of traffic and transportation engineering (English edition),* vol. 3, no. 1, pp. 51-58, 2016.

[3] C. Zhang, J. Hu, J. Qiu, W. Yang, H. Sun, and Q. Chen, "A novel fuzzy observer-based steering control approach for path tracking in autonomous vehicles," *IEEE Transactions on Fuzzy Systems,* vol. 27, no. 2, pp. 278-290, 2018.

[4] Tesla, "Tesla Vehicle Safety Report," Tesla, TX, Austin, USA, 2023.

[5] S. Feng, X. Yan, H. Sun, Y. Feng, and H. X. Liu, "Intelligent driving intelligence test for autonomous vehicles with naturalistic and adversarial environment," *Nature communications,* vol. 12, no. 1, p. 748, 2021.

[6] E. Lee, Y. Han, J.-Y. Lee, and B. Son, "Modeling Lane-Changing Behaviors for Autonomous Vehicles based on Vehicle-to-Vehicle Communication," *IEEE Access,* 2023.

[7] Z. Zheng, "Recent developments and research needs in modeling lane changing," *Transportation research part B: methodological,* vol. 60, pp. 16-32, 2014.

[8] Y. Zhang, Q. Xu, J. Wang, K. Wu, Z. Zheng, and K. Lu, "A learning-based discretionary lane-change decision-making model with driving style awareness," *IEEE transactions on intelligent transportation systems,* vol. 24, no. 1, pp. 68-78, 2022.

[9] B. Khelfa, I. Ba, and A. Tordeux, "Predicting highway lane-changing maneuvers: A benchmark analysis of machine and ensemble learning algorithms," *Physica A: Statistical Mechanics and its Applications,* vol. 612, p. 128471, 2023.

[10] A. Kusuma, R. Liu, C. Choudhury, and F. Montgomery, "Analysis of the driving behaviour at weaving section using multiple traffic surveillance data," *Transportation Research Procedia,* vol. 3, pp. 51-59, 2014.

[11] B. Son, T. Kim, H. J. Kim, and S. Lee, "Probabilistic model of traffic breakdown with random propagation of disturbance for ITS application," in *Knowledge-Based Intelligent Information and Engineering Systems: 8th International Conference, KES 2004, Wellington, New Zealand, September 20-25, 2004, Proceedings, Part III 8,* 2004: Springer, pp. 45-51.

[12] Y. Yu, X. Luo, Q. Su, and W. Peng, "A dynamic lane-changing decision and trajectory planning model of autonomous vehicles under mixed autonomous vehicle and human-driven vehicle environment," *Physica A: Statistical Mechanics and its Applications,* vol. 609, p. 128361, 2023.

[13] X. Xiao, P. Bo, Y. Chen, Y. Chen, and K. Li, "Enhancing lane changing trajectory prediction on highways: A heuristic attention-based encoder-decoder model," *Physica A: Statistical Mechanics and its Applications,* p. 129696, 2024.

[14] Y. Atzmon, F. Kreuk, U. Shalit, and G. Chechik, "A causal view of compositional zero-shot recognition," *Advances in Neural Information Processing Systems,* vol. 33, pp. 1462-1473, 2020.

[15] T. Misu and Y.-T. Chen, "Toward reasoning of driving behavior," in *2018 21st International Conference on Intelligent Transportation Systems (ITSC),* 2018: IEEE, pp. 204-209.

[16] M. M. R. Komol *et al.,* "Deep RNN based prediction of driver's intended movements at intersection using cooperative awareness messages," *IEEE Transactions on Intelligent Transportation Systems,* 2023.

[17] S. Noh, "Decision-making framework for autonomous driving at road intersections: Safeguarding against collision, overly conservative behavior, and violation vehicles," *IEEE Transactions on Industrial Electronics,* vol. 66, no. 4, pp. 3275-3286, 2018.

[18] P. S. Chib and P. Singh, "Recent advancements in end-to-end autonomous driving using deep learning: A survey," *IEEE Transactions on Intelligent Vehicles,* 2023.

[19] X. Yan, Z. Zou, S. Feng, H. Zhu, H. Sun, and H. X. Liu, "Learning naturalistic driving environment with statistical realism," *Nature communications,* vol. 14, no. 1, p. 2037, 2023.

[20] N. Phuksaksakul, S. Yasmin, and M. M. Haque, "A random parameters copula-based binary logit-generalized ordered logit model with parameterized dependency: application to active traveler injury severity analysis," *Analytic methods in accident research,* vol. 38, p. 100266, 2023.

[21] M. M. R. Komol, M. M. Hasan, M. Elhenawy, S. Yasmin, M. Masoud, and A. Rakotonirainy, "Crash severity analysis of vulnerable road users using machine learning," *PLoS one,* vol. 16, no. 8, p. e0255828, 2021.

[22] L. Evans, "Driver injury and fatality risk in two-car crashes versus mass ratio inferred using Newtonian mechanics," *Accident Analysis & Prevention,* vol. 26, no. 5, pp. 609-616, 1994.

[23] U. Björnstig, J. Björnstig, and A. Eriksson, "Passenger car collision fatalities–with special emphasis on collisions with heavy vehicles," *Accident analysis & prevention,* vol. 40, no. 1, pp. 158-166, 2008.

[24] S. Lyman and E. R. Braver, "Occupant deaths in large truck crashes in the United States: 25 years of experience," *Accident Analysis & Prevention,* vol. 35, no. 5, pp. 731-739, 2003.

[25] W. Zou, X. Wang, and D. Zhang, "Truck crash severity in New York city: an investigation of the spatial and the time of day effects," *Accident Analysis & Prevention,* vol. 99, pp. 249-261, 2017.

[26] Y. Wang *et al.,* "Decision-making driven by driver intelligence and environment reasoning for high-level autonomous vehicles: a survey," *IEEE Transactions on Intelligent Transportation Systems,* 2023.

[27] D. Krajzewicz, G. Hertkorn, C. Rössel, and P. Wagner, "SUMO (Simulation of Urban MObility)-an open-source traffic simulation," in *Proceedings of the 4th middle East Symposium on Simulation and Modelling (MESM20002),* 2002, pp. 183-187.

[28] R. Krajewski, J. Bock, L. Klocker, and L. Eckstein, "The highd dataset: A drone dataset of naturalistic vehicle trajectories on german highways for validation of highly automated driving systems," in *2018 21st international conference on intelligent transportation systems (ITSC),* 2018: IEEE, pp. 2118-2125.

[29] H. Zhou and J. Laval, "Longitudinal motion planning for autonomous vehicles and its impact on congestion: A survey," 2019.

[30] X. Wang, R. Jiang, L. Li, Y. Lin, X. Zheng, and F.-Y. Wang, "Capturing car-following behaviors by deep learning," *IEEE Transactions on Intelligent Transportation Systems,* vol. 19, no. 3, pp. 910-920, 2017.

[31] M. Zhu, X. Wang, and Y. Wang, "Human-like autonomous car-following model with deep reinforcement learning," *Transportation research part C: emerging technologies,* vol. 97, pp. 348-368, 2018.

[32] J. Wang, L. Zhang, D. Zhang, and K. Li, "An adaptive longitudinal driving assistance system based on driver characteristics," *IEEE Transactions on Intelligent Transportation Systems,* vol. 14, no. 1, pp. 1-12, 2012.

[33] Z. Huang, X. Xu, H. He, J. Tan, and Z. Sun, "Parameterized batch reinforcement learning for longitudinal control of autonomous land

vehicles," *IEEE Transactions on Systems, Man, and Cybernetics: Systems,* vol. 49, no. 4, pp. 730-741, 2017.

[34] I. Miller *et al.*, "Team Cornell's Skynet: Robust perception and planning in an urban environment," *Journal of Field Robotics,* vol. 25, no. 8, pp. 493-527, 2008.

[35] J.-H. Ryu, D. Ogay, S. Bulavintsev, H. Kim, and J.-S. Park, "Development and experiences of an autonomous vehicle for high-speed navigation and obstacle avoidance," in *Frontiers of Intelligent Autonomous Systems*: Springer, 2013, pp. 105-116.

[36] H. Wang, Y. Huang, A. Khajepour, Y. Zhang, Y. Rasekhipour, and D. Cao, "Crash mitigation in motion planning for autonomous vehicles," *IEEE transactions on intelligent transportation systems,* vol. 20, no. 9, pp. 3313-3323, 2019.

[37] X. Zeng and J. Wang, "Globally energy-optimal speed planning for road vehicles on a given route," *Transportation Research Part C: Emerging Technologies,* vol. 93, pp. 148-160, 2018.

[38] Y. Lin, J. McPhee, and N. L. Azad, "Comparison of deep reinforcement learning and model predictive control for adaptive cruise control," *IEEE Transactions on Intelligent Vehicles,* vol. 6, no. 2, pp. 221-231, 2020.

[39] P. Wang, C.-Y. Chan, and A. de La Fortelle, "A reinforcement learning based approach for automated lane change maneuvers," in *2018 IEEE Intelligent Vehicles Symposium (IV)*, 2018: IEEE, pp. 1379-1384.

[40] Y. Chen, C. Dong, P. Palanisamy, P. Mudalige, K. Muelling, and J. M. Dolan, "Attention-based hierarchical deep reinforcement learning for lane change behaviors in autonomous driving," in *Proceedings of the IEEE/CVF Conference on Computer Vision and Pattern Recognition Workshops*, 2019, pp. 0-0.

[41] V. Rausch, A. Hansen, E. Solowjow, C. Liu, E. Kreuzer, and J. K. Hedrick, "Learning a deep neural net policy for end-to-end control of autonomous vehicles," in *2017 American Control Conference (ACC)*, 2017: IEEE, pp. 4914-4919.

[42] D. Chen, M. Zhu, H. Yang, X. Wang, and Y. Wang, "Data-driven Traffic Simulation: A Comprehensive Review," *IEEE Transactions on Intelligent Vehicles,* 2024.

[43] B. R. Kiran *et al.*, "Deep reinforcement learning for autonomous driving: A survey," *IEEE Transactions on Intelligent Transportation Systems,* 2021.

[44] O. Vinyals *et al.*, "Grandmaster level in StarCraft II using multi-agent reinforcement learning," *Nature,* vol. 575, no. 7782, pp. 350-354, 2019.

[45] Y. Du, J. Chen, C. Zhao, C. Liu, F. Liao, and C.-Y. Chan, "Comfortable and energy-efficient speed control of autonomous vehicles on rough pavements using deep reinforcement learning," *Transportation Research Part C: Emerging Technologies,* vol. 134, p. 103489, 2022.

[46] J. Liao, T. Liu, X. Tang, X. Mu, B. Huang, and D. Cao, "Decision-making strategy on highway for autonomous vehicles using deep reinforcement learning," *IEEE Access,* vol. 8, pp. 177804-177814, 2020.

[47] A. J. M. Muzahid, S. F. Kamarulzaman, and M. A. Rahman, "Comparison of ppo and sac algorithms towards decision making strategies for collision avoidance among multiple autonomous vehicles," in *2021 International Conference on Software Engineering & Computer Systems and 4th International Conference on Computational Science and Information Management (ICSECS-ICOCSIM)*, 2021: IEEE, pp. 200-205.

[48] S. Han and F. Miao, "Behavior planning for connected autonomous vehicles using feedback deep reinforcement learning," *arXiv preprint arXiv:2003.04371*, 2020.

[49] S. Nageshrao, H. E. Tseng, and D. Filev, "Autonomous highway driving using deep reinforcement learning," in *2019 IEEE International Conference on Systems, Man and Cybernetics (SMC)*, 2019: IEEE, pp. 2326-2331.

[50] G. Li, Y. Yang, S. Li, X. Qu, N. Lyu, and S. E. Li, "Decision making of autonomous vehicles in lane change scenarios: Deep reinforcement learning approaches with risk awareness," *Transportation research part C: emerging technologies,* vol. 134, p. 103452, 2022.

[51] X. Ma, L. Xia, Z. Zhou, J. Yang, and Q. Zhao, "Dsac: Distributional soft actor critic for risk-sensitive reinforcement learning," *arXiv preprint arXiv:2004.14547*, 2020.

[52] G. Du, Y. Zou, X. Zhang, Z. Li, and Q. Liu, "Hierarchical motion planning and tracking for autonomous vehicles using global heuristic based potential field and reinforcement learning based predictive control," *IEEE Transactions on Intelligent Transportation Systems,* 2023.

[53] M. Al-Sharman, R. Dempster, M. A. Daoud, M. Nasr, D. Rayside, and W. Melek, "Self-learned autonomous driving at unsignalized intersections: A hierarchical reinforced learning approach for feasible decision-making," *IEEE Transactions on Intelligent Transportation Systems,* 2023.

[54] Q. Cui, R. Ding, C. Wei, and B. Zhou, "A hierarchical framework of emergency collision avoidance amid surrounding vehicles in highway driving," *Control Engineering Practice,* vol. 109, p. 104751, 2021.

[55] Y. Song, A. Romero, M. Müller, V. Koltun, and D. Scaramuzza, "Reaching the limit in autonomous racing: Optimal control versus reinforcement learning," *Science Robotics,* vol. 8, no. 82, p. eadg1462, 2023.

[56] J. Kim and D. Kum, "Collision risk assessment algorithm via lane-based probabilistic motion prediction of surrounding vehicles," *IEEE Transactions on Intelligent Transportation Systems,* vol. 19, no. 9, pp. 2965-2976, 2017.

[57] J. Hillenbrand, A. M. Spieker, and K. Kroschel, "A multilevel collision mitigation approach—Its situation assessment, decision making, and performance tradeoffs," *IEEE Transactions on intelligent transportation systems,* vol. 7, no. 4, pp. 528-540, 2006.

[58] C. Chen, X. Liu, H.-H. Chen, M. Li, and L. Zhao, "A rear-end collision risk evaluation and control scheme using a Bayesian network model," *IEEE Transactions on Intelligent Transportation Systems,* vol. 20, no. 1, pp. 264-284, 2018.

[59] J. E. Naranjo, C. Gonzalez, R. Garcia, and T. De Pedro, "Lane-change fuzzy control in autonomous vehicles for the overtaking maneuver," *IEEE Transactions on Intelligent Transportation Systems,* vol. 9, no. 3, pp. 438-450, 2008.

[60] S. Glaser, B. Vanholme, S. Mammar, D. Gruyer, and L. Nouveliere, "Maneuver-based trajectory planning for highly autonomous vehicles on real road with traffic and driver interaction," *IEEE Transactions on intelligent transportation systems,* vol. 11, no. 3, pp. 589-606, 2010.

[61] J.-H. Kim and D.-S. Kum, "Threat prediction algorithm based on local path candidates and surrounding vehicle trajectory predictions for automated driving vehicles," in *2015 IEEE Intelligent Vehicles Symposium (IV)*, 2015: IEEE, pp. 1220-1225.

[62] K. Lee and D. Kum, "Collision avoidance/mitigation system: Motion planning of autonomous vehicle via predictive occupancy map," *IEEE Access,* vol. 7, pp. 52846-52857, 2019.

[63] L. Li, J. Gan, X. Ji, X. Qu, and B. Ran, "Dynamic driving risk potential field model under the connected and automated vehicles environment and its application in car-following modeling," *IEEE Transactions on Intelligent Transportation Systems,* vol. 23, no. 1, pp. 122-141, 2020.

[64] D. Ni, "A unified perspective on traffic flow theory, part I: the field theory," in *Icctp 2011: Towards sustainable transportation systems*, 2011, pp. 4227-4243.

[65] K. Mokhtari and A. R. Wagner, "Don't Get Yourself into Trouble! Risk-aware Decision-Making for Autonomous Vehicles," *arXiv preprint arXiv:2106.04625*, 2021.

[66] R. S. Sutton, D. McAllester, S. Singh, and Y. Mansour, "Policy gradient methods for reinforcement learning with function approximation," *Advances in neural information processing systems,* vol. 12, 1999.

[67] Z. Fan, R. Su, W. Zhang, and Y. Yu, "Hybrid actor-critic reinforcement learning in parameterized action space," *arXiv preprint arXiv:1903.01344*, 2019.

[68] J. Schulman, F. Wolski, P. Dhariwal, A. Radford, and O. Klimov, "Proximal policy optimization algorithms," *arXiv preprint arXiv:1707.06347*, 2017.

[69] J. Schulman, P. Moritz, S. Levine, M. Jordan, and P. Abbeel, "High-dimensional continuous control using generalized advantage estimation," *arXiv preprint arXiv:1506.02438*, 2015.

[70] A. Vaswani *et al.*, "Attention is all you need," *Advances in neural information processing systems,* vol. 30, 2017.

[71] Y. Wang, H. Tu, N. Sze, H. Li, and X. Ruan, "A novel traffic conflict risk measure considering the effect of vehicle weight," *Journal of safety research,* vol. 80, pp. 1-13, 2022.

[72] K. Cho, T. Ha, G. Lee, and S. Oh, "Deep predictive autonomous driving using multi-agent joint trajectory prediction and traffic rules," in *2019 IEEE/RSJ International Conference on Intelligent Robots and Systems (IROS)*, 2019: IEEE, pp. 2076-2081.

[73] M. Fu, T. Zhang, W. Song, Y. Yang, and M. Wang, "Trajectory prediction-based local spatio-temporal navigation map for autonomous driving in dynamic highway environments," *IEEE Transactions on Intelligent Transportation Systems,* vol. 23, no. 7, pp. 6418-6429, 2021.

[74] X. Wen, S. Jian, and D. He, "Modeling the effects of autonomous vehicles on human driver car-following behaviors using inverse reinforcement learning," *IEEE Transactions on Intelligent Transportation Systems,* 2023.

[75] D. K. Patel, D. Singh, and B. Singh, "A comparative analysis for impact of distributed generations with electric vehicles planning," *Sustainable Energy Technologies and Assessments,* vol. 52, p. 101840, 2022.

[76] D. K. Patel, D. Singh, and B. Singh, "Impact assessment of distributed generations with electric vehicles planning: A review," *Journal of Energy Storage,* vol. 43, p. 103092, 2021.

[77] B. Ma *et al.*, "Deep reinforcement learning of UAV tracking control under wind disturbances environments," *IEEE Transactions on Instrumentation and Measurement,* 2023.

[78] I. Loshchilov and F. Hutter, "Decoupled weight decay regularization," *arXiv preprint arXiv:1711.05101,* 2017.

[79] J. Erdmann, "SUMO's lane-changing model," in *Modeling Mobility with Open Data: 2nd SUMO Conference 2014 Berlin, Germany, May 15-16, 2014,* 2015: Springer, pp. 105-123.

[80] M. A. Silgu, İ. G. Erdağı, G. Göksu, and H. B. Celikoglu, "Combined control of freeway traffic involving cooperative adaptive cruise controlled and human driven vehicles using feedback control through SUMO," *IEEE Transactions on Intelligent Transportation Systems,* vol. 23, no. 8, pp. 11011-11025, 2021.

[81] J. C. Hayward, "Near miss determination through use of a scale of danger," 1972.

[82] D. F. Cooper and N. Ferguson, "Traffic studies at T-Junctions. 2. A conflict simulation Record," *Traffic Engineering & Control,* vol. 17, no. Analytic, 1976.

[83] P. Cooper, "Experience with traffic conflicts in Canada with emphasis on "post encroachment time" techniques," in *International calibration study of traffic conflict techniques,* 1984: Springer, pp. 75-96.

[84] P. A. Lopez *et al.*, "Microscopic traffic simulation using sumo," in *2018 21st international conference on intelligent transportation systems (ITSC),* 2018: IEEE, pp. 2575-2582.

[85] A. Dijkstra and H. Drolenga, "Safety effects of route choice in a road network: Simulation of changing route choice," SWOV Institute for Road Safety Research, 2008.

[86] K. Yang, X. Tang, S. Qiu, S. Jin, Z. Wei, and H. Wang, "Towards robust decision-making for autonomous driving on highway," *IEEE Transactions on Vehicular Technology,* 2023.

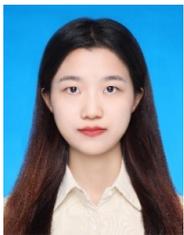
**Di Chen** received the B.Sc. and M.S. degree in Transportation Engineering from School of Transportation engineering, Tongji University, Shanghai, China. Currently, she is a research assistant in the Thrust of Intelligent Transportation under the Systems Hub at the Hong Kong University of Science and Technology (Guangzhou). Her research interests include traffic simulation, machine learning methods for autonomous vehicles.

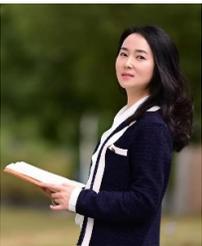
**Hao Li** received the B.Sc. degree in civil engineering from Tongji University, Shanghai, China, and the M.Sc. and Ph.D. degrees in transportation and planning from Delft University of Technology, Delft, The Netherlands. Dr. Li is a Professor with the College of Transportation Engineering, Tongji University. Her main research interests include transport network modeling, travel behavior, and automatic vehicle driving and road testing. She has hosted three NSFC projects, one National Key R&D Program of China, two research projects funded by Ministry of Education (MoE), etc. She has (co)authored more than 60 scientific papers, including Transportation Research Part A, IEEE Transactions on ITS, etc.

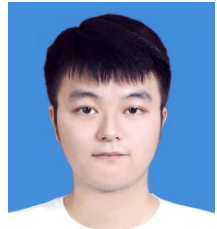
**Zhicheng Jin** received the B.Sc. degree in Transportation from School of Transportation and Logistics, Southwest Jiaotong University, and is currently working toward M.S. degree at School of Transportation engineering, Tongji University. His research interest includes public transportation planning and operation.

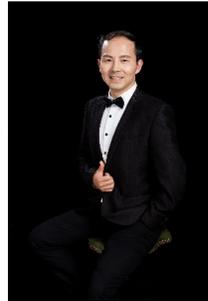
**Huizhao Tu** received the B.Sc. degree in civil engineering and the M.Sc. degree in transportation engineering from Tongji University, China, the Ph.D. degree in Transport & Planning from Delft University of Technology, The Netherlands. He is currently a full professor with College of Transportation Engineering and Vice Dean of Institute for Urban Risk Management at Tongji University, Vice Chairman of Shanghai Standard Technical Committees in Intelligent Transportation System, China. He was with Department of Transport & Planning, Delft University of Technology, The Netherlands, where he was an Assistant Researcher from 2003 to 2008.

His research interests include autonomous vehicles, transport risk assessment, smart transportation, data fusion etc. From 2003 to 2010, he participated in several Dutch projects on reliable traffic monitoring and emergency traffic management, and from 2010 till now, he was the PIs in several Chinese projects (e.g. NSFC, NSFC-NWO, MOST) in autonomous vehicles, transport risk assessment and smart transportation. Dr. Tu has published over 100 peer reviewed papers, three books and two Shanghai Standards. One of the books is Risk Management of Autonomous Vehicles Road Testing (ISBN 978-7-5765-0094-3). He got five Municipality Awards in Science and Technology Progressing.

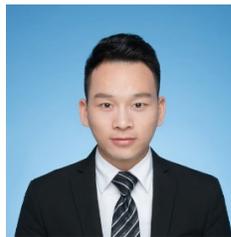
**Meixin Zhu** is a tenure-track Assistant Professor in the Thrust of Intelligent Transportation (INTR) under the Systems Hub at the Hong Kong University of Science and Technology (Guangzhou) and an affiliated Assistant Professor in the Civil and Environmental Engineering Department at the Hong Kong University of Science and Technology. He is also with Guangdong Provincial Key Lab of Integrated Communication, Sensing and Computation for Ubiquitous Internet of Things. He obtained a Ph.D. degree in intelligent transportation at the University of Washington (UW) in 2022. He received his BS and MS degrees in traffic engineering in 2015 and 2018, respectively, from Tongji University. His research interests include Autonomous Driving Decision Making and Planning, Driving Behavior Modeling, Traffic-Flow Modeling and Simulation, Traffic Signal Control, and (Multi-Agent) Reinforcement Learning. He is a recipient of the TRB Best Dissertation Award (AED50) in 2023.